\documentclass[a4paper,twoside]{article}

\usepackage{epsfig}
\usepackage{subfigure}
\usepackage{calc}
\usepackage{amssymb}
\usepackage{amstext}
\usepackage{amsmath}
\usepackage{amsthm}
\usepackage{multicol}
\usepackage{pslatex}
\usepackage{apalike}
\usepackage{graphicx}
\usepackage{SCITEPRESS}  
\usepackage{xcolor,colortbl}
\definecolor{stacking}{RGB}{217, 230, 242}
\definecolor{RF}{RGB}{32, 223, 32}
\definecolor{LR}{RGB}{249, 55, 6}
\usepackage{floatrow}
\newfloatcommand{capbtabbox}{table}[\captop][\FBwidth]

\begin{document}

\title{Demand Prediction Using Machine Learning Methods and Stacked Generalization}

\author{\authorname{Resul Tugay, \c{S}ule G\"{u}nd\"{u}z \"{O}\c{g}\"{u}d\"{u}c\"{u}}
\affiliation{Department of Computer Engineering, Istanbul Technical University,Istanbul, Turkey}
\email{\{tugayr, sgunduz\}@itu.edu.tr}
}


\abstract{Supply and demand are two fundamental concepts of sellers and customers. Predicting demand accurately is critical for organizations in order to be able to make plans. In this paper, we propose a new approach for demand prediction on an e-commerce web site. The proposed model differs from earlier models in several ways. The business model used in the e-commerce web site, for which the model is implemented, includes many sellers that sell the same product at the same time at different prices where the company operates a market place model. The demand prediction for such a model should consider the price of the same product sold by competing sellers along the features of these sellers.
 In this study we first applied different regression algorithms for specific set of products of one department of a company that is one of the most popular online e-commerce companies in Turkey. Then we used stacked generalization or also known as stacking ensemble learning to predict demand. Finally, all the approaches are evaluated on a real world data set obtained from the e-commerce company. The experimental results show that some of the machine learning methods do produce almost as good results as the stacked generalization method.}

\onecolumn \maketitle \normalsize \vfill

\section{\uppercase{Introduction}}
\label{sec:introduction}

\noindent 	Demand forecasting is the concept of predicting the quantity of a product that consumers will purchase during a specific time period.	 Predicting right demand of a product is an important phenomenon in terms of space, time and money for the sellers.  Sellers may have limited time or need to sell their products as soon as possible due to the storage and money restrictions. Therefore demand of a product depends on many factors such as price, popularity, time, space etc. Forecasting demand is being hard when the number of factors increases. Demand prediction is also closely related with seller revenue. If sellers store much more product than the demand then this may lead to surplus \cite{miller1988effects}. On the other hand storing less product in order to save inventory costs when the product has high demand will cause less revenue. Because of these and many more reasons, demand forecasting has become an interesting and important topic for researchers in many areas
such as water demand prediction \cite{an1996discovering}, data center application \cite{gmach2007workload} and energy demand prediction \cite{srinivasan2008energy}.
		

\begin{figure*}
  \includegraphics[width=\textwidth]{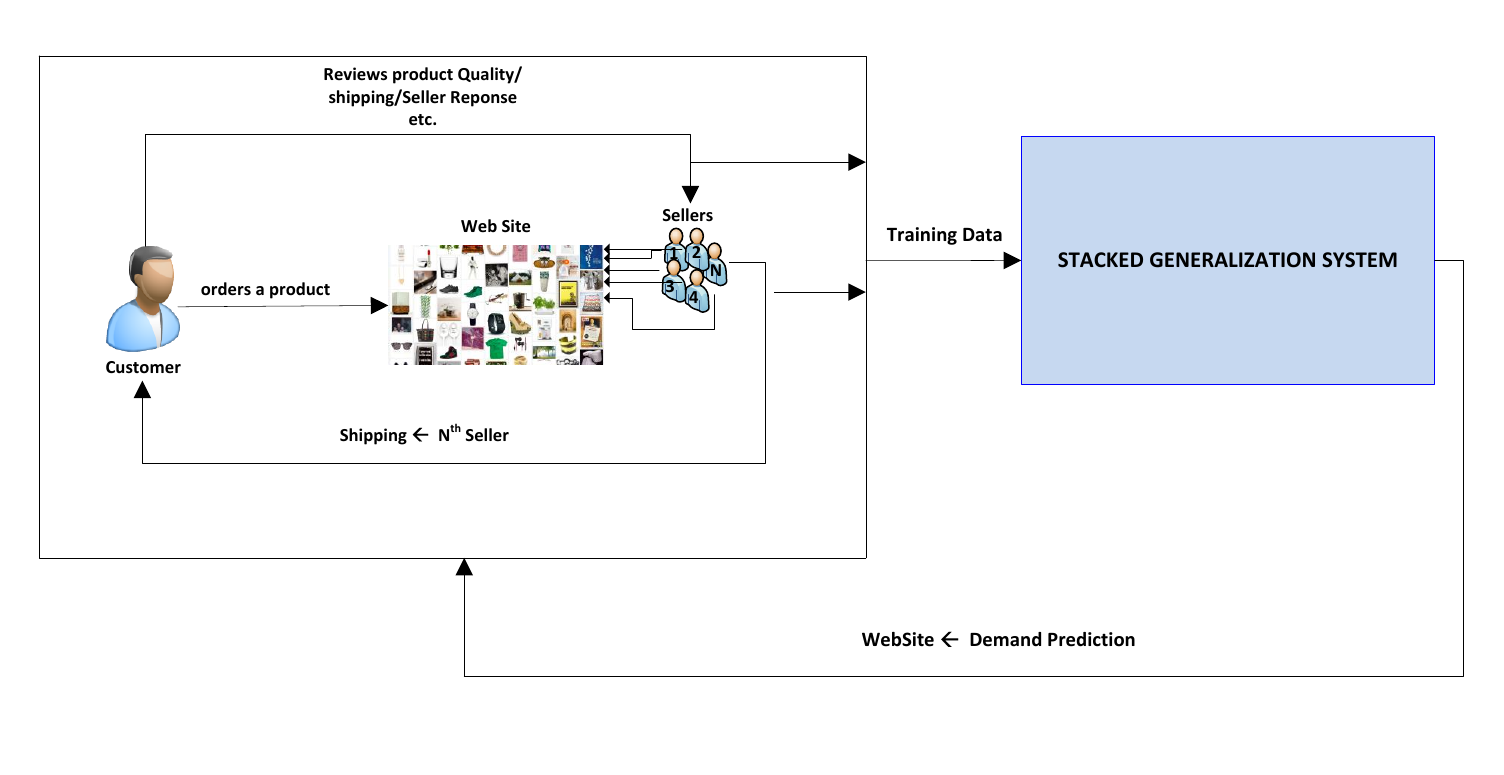}
  \caption{General System Schema}
   \label{fig:fig1}
\end{figure*}
	 
	 The rest of the paper is organized as follows:Section 2 discusses related work and section 3 describe methodology. In section 5 we describe our experimental results and data definitions. In section 6 we conclude this work and future work.

	\section{RELATED WORK}
\noindent In literature, the research studies on demand forecasting can be grouped into three main categories: (1) Statistical Methods; (2) Artificial Intelligence Methods; (3) Hybrid Methods.
		 \newline
	\textbf{\emph{Statistical Methods:}} Linear regression, regression tree, moving average, weighted average, bayesian analysis are just some of statistical methods for demand forecasting \cite{liu2013sales}. Johnson et al. used regression trees to predict demand due to its simplicity and interpretability \cite{johnson2014analytics}. They applied demand prediction on data given by an online retailer web company named as Rue La La. The web site consists of several events which are changing within 1-4 days interval from different departments. Each event has multiple products called "style" and each product has different items. Items are typical products that have different properties such as size and color. Because of the price is set at style level, they aggregate items at style level and use different regression models for each department. Ediger and Akar used Autoregressive Integrated Moving Average (ARIMA) and seasonal ARIMA (SARIMA) methods to predict future energy demand of Turkey from 2005 to 2020 \cite{ediger2007arima}. Also Lim and McAleer used ARIMA for travel demand forecasting \cite{lim2002time}. \newline
	Although basic implementation and simple interpretation of statistical methods, different approaches are applied for demand forecasting such as Artificial Intelligence (AI) and hybrid methods.\newline
	\textbf{\emph{AI Methods:}} AI methods are commonly used in literature for demand forecasting due to their primary advantage of being efficient and accurate \cite{chang2006fuzzy}, \cite{gutierrez2008lumpy}, \cite{zhang1998forecasting}, \cite{yoo1999short}. Frank et al. used Artificial Neural Networks (ANN) to predict women's apparel sales and ANN outperformed two statistical based  models \cite{frank2003forecasting}. Sun et al. proposed a novel extreme learning machine which is a type of neural network for sales forecasting. The proposed methods outperformed traditional neural networks for sales forecasting \cite{sun2008sales}.\newline
	\textbf{\emph{Hybrid Methods:}} Another method of forecasting sales or demands is hybrid methods. Hybrid methods utilize more than one method and use the strength of these methods. Zhang used ARIMA and ANN hybrid methodology in time series forecasting and proposed a method that achieved more accuracy than the methods when they were used separately \cite{zhang2003time}. In addition to hybrid models, there has been some studies where fuzzy logic is used for demand forecasting \cite{aburto2007improved}, \cite{thomassey2002short}, \cite{vroman1998fuzzy}.
		
Thus far, many researchers focused on different statistical, AI and hybrid methods for forecasting problem. But Islek and Gunduz Oguducu proposed a state-of-art method that is based on stack generalization on the problem of forecasting demand of warehouses and their model decreased the error rate using proposed method \cite{islek}. On the contrary, in this paper we use different statistical methods and compare their results with stack generalization method which uses these methods as sub-level learners.
\section{METHODOLOGY}
\noindent This section describes the methodology and techniques used to solve demand prediction problem.
		
	\subsection{Stacked Generalization}
\begin{figure}[h!]
	\centering
 	 \includegraphics[width=\linewidth]			{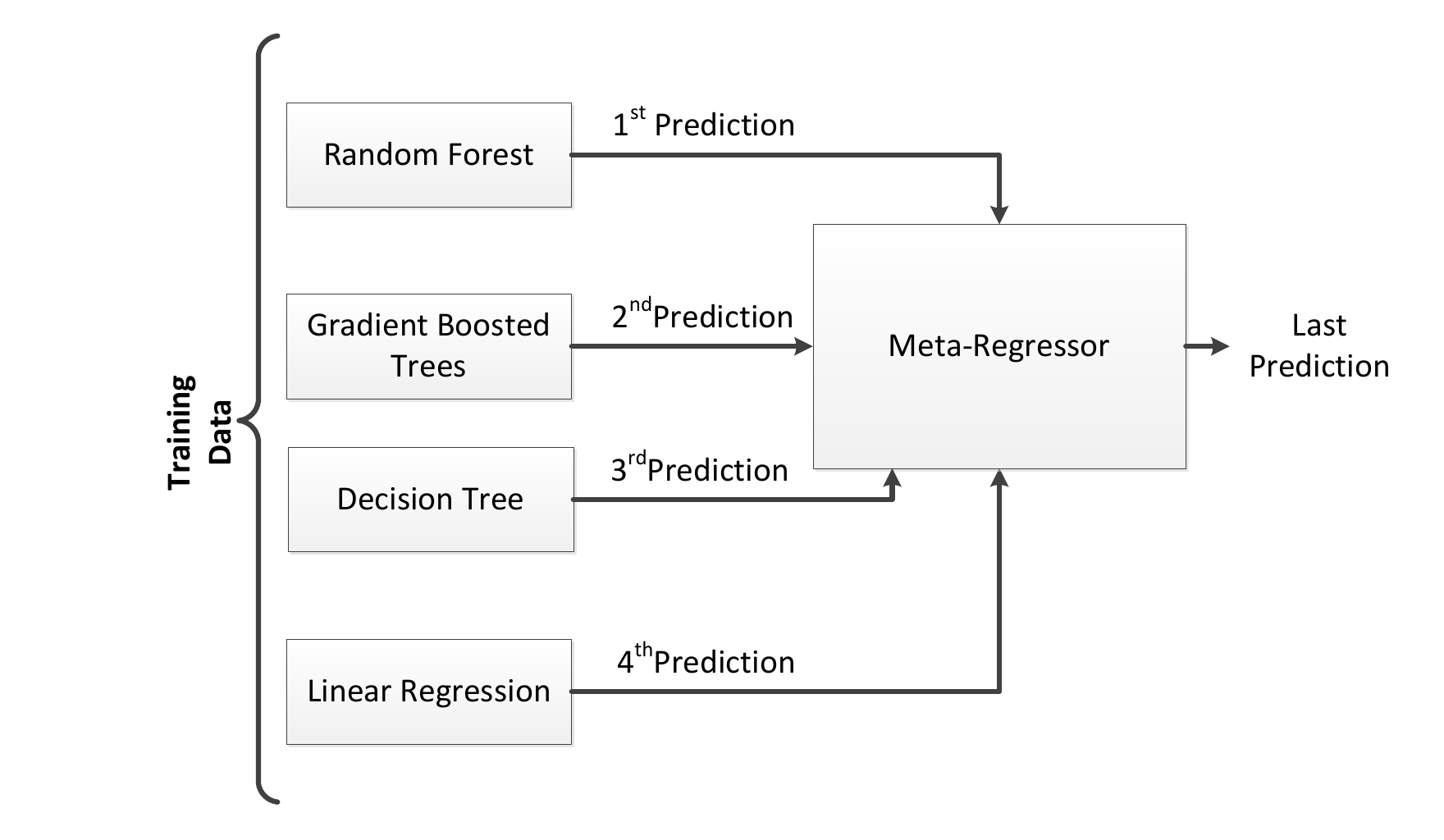}
	  \caption{Stacked Generalization}
 	 \label{fig:fig2}
	\end{figure}

\noindent Stacked generalization (SG) is one of the ensemble methods applied in machine learning which use multiple learning algorithms to improve the predictive performance. It is based on training a learning algorithm to combine the predictions of the learning algorithms involved instead of selecting a single learning algorithm. Although there are many different ways to implement stacked generalization, its primary implementation consists of two stages. At the first stage, all the learning algorithms are trained using the available data. At this stage, we use linear regression, random forest regression, gradient boosting and decision tree regression as the first-level regressors. At the second stage a combiner algorithm is used to make a final prediction based on all the predictions of the learning algorithms applied in the first stage.  At this stage, we use again same regression algorithms we used in the first stage to specify which model is the best regressor for this problem. The general schema of stacked generalization applied in this study can be seen in Figure \ref{fig:fig2}.
	 
  \subsection{Linear Regression}
A linear regression (LR) is a statistical method where a dependent variable $\gamma$ (the target variable) is computed from $p$ independent variables that are assumed to have an influence on the target variable. Given a data set  of $n$ data points, the formula for a regression of one data point $\gamma_i$ (regressand) is as follows:
  \begin{equation} \label{eq:linear_regression}
    \gamma_i = \beta_jx_{i1} +..\beta_px_{ip} + \epsilon_i \enspace \enspace \enspace i = 1,2,..n
   \end{equation}
where ${\beta_{j}}$ is the regression coefficient that can be calculated using Least Squares  approach,  $x_{ij}$ (regressor) is the value of the $j^{th}$ independent variable and $\epsilon_i$ the error term. 
The best-fitting straight line for the observed data is calculated by minimizing the loss function which is sum of the squares of differences between the value of the point $\gamma_i$ and the predicted value $\hat{\gamma_i}$ (the value on the line) as shown in Equation \ref{eq:mse}. 
\begin{equation} \label{eq:mse}
	MSE = \frac{1}{n}\sum_{i=1}^{n} (\hat{\gamma_i}- \gamma_i)^2
\end{equation}
The best values of regression coefficients and the error terms can be found by minimizing the loss function in Equation \ref{eq:mse}. While minimizing loss function, a penalty term is used to control the complexity of the model. For instance, lasso (least absolute shrinkage and selection operator) uses $L1$ norm penalty term and ridge regression uses  $L2$ norm penalty term as shown in Equation \eqref{eq:lasso} and \eqref{eq:ridge} respectively. $\lambda$ is regularization parameter that prevents overfitting or controls model complexity. In both Equation \eqref{eq:lasso} and \eqref{eq:ridge} coefficients ($\beta$), dependent variables ($\gamma$) and independent variables ($X$) are represented in matrix form.
\begin{equation} \label{eq:lasso}
	\beta = argmin \lbrace \frac{1}{n}(\gamma-\beta X)^2 + \lambda_1||\beta||_1 \rbrace
\end{equation}
\begin{equation} \label{eq:ridge}
	\beta = argmin \lbrace \frac{1}{n}(\gamma-\beta X)^2 + \lambda_2||\beta||_2^2 \rbrace
\end{equation}
In this study, we used elastic net which is combination of $L1$ and $L2$ penalty terms with $\lambda = 0.8$ for $L1$ and $\lambda = 0.2$ for $L2$ penalty term with $\lambda = 0.3$ regularization parameter as shown in Equation \eqref{eq:elas}.
\begin{equation} \label{eq:elas}
	\beta = argmin \lbrace \frac{1}{n}(\gamma-\beta X)^2 + \lambda(0.8||\beta||_1 + 0.2||\beta||_2^2) \rbrace
\end{equation}

 \subsection{Decision Tree Regression}
Decision trees (DT) can be used both in classification and regression problems. Quinlan proposed ID3 algorithm as the first decision tree algorithm \cite{quinlan1986induction}. Decision tree algorithms classify both categorical (classification) and numerical (regression) samples in a form of tree structure with a root node, internal nodes and leaf nodes. Internal nodes contain one of the possible input variables (features) available at that point in the tree. The selection of input variable is chosen using information gain or impurity for classification problems and standard deviation reduction for regression problems. The leaves represent labels/predictions. Random forest and gradient boosting algorithms are both decision tree based algorithms.
In this study, decision tree method is applied for regression problems where variance reduction is employed for selection of variables in the internal nodes. Firstly, variance of root node is calculated using Equation \ref{eq:stdroot}, then variance of features is calculated using Equation \ref{eq:stdatt} to construct the tree.
\begin{equation} \label{eq:stdroot}
	\sigma^2 = \frac{\sum_{i=1}^{n}({x_i - \mu})^2}{n}
\end{equation}
In Equation \ref{eq:stdroot}, $n$ is the total number of samples and $\mu$ is  the mean of the samples in the training set. After calculating variance of the root node, variance of input variables is calculated as follows:
\begin{equation} \label{eq:stdatt}
	\sigma^2_{X} = \sum_{c \epsilon X}^{ }{P(c)\sigma^2_{c}}
\end{equation}
 In Equation \ref{eq:stdatt}, $X$ is the input variable and $c$'s are the distinct values of this feature. For example, $X :$ Brand and $c :$ Samsung, Apple or Nokia. $P(c)$ is the probability of $c$ being in the attribute $X$ and $\sigma^2_{c}$ is the variance of the value $c$. Input variable that has the minimum variance or largest variance reduction is selected as the best node as shown in Equation \ref{eq:str}:
\begin{equation} \label{eq:str}
	vr_{X} = \sigma^2 - \sigma^2_{X}
\end{equation}
Finally leaves are representing the average values of instances that they include in subsection 3.4 with bootstrapping method. This process continues recursively, until variance of leaves is smaller than a threshold or all input variables are used. Once a tree has been constructed, new instance is tested by asking questions to the nodes in the tree. When reaching a leaf, value of that leaf is taken as prediction.

\subsection{Random Forest}
	Random forest (RF) is a type of meta learner that uses number of decision trees for both classification and regression problems \cite{breiman2001random}. The features and samples are drawn randomly for every tree in the forest and these trees are trained independently. Each tree is constructed with bootstrap sampling method. Bootstrapping relies on sampling with replacement. Given a dataset $D$ with $N$ samples, a training data set of size $N$ is created by sampling from $D$ with replacement. The remaining samples in $D$ that are not in the training set are separated as the test set. This kind of sampling is called bootstrap sampling. 

\begin{figure}[h!]
	\centering
 	 \includegraphics[width=\linewidth]			{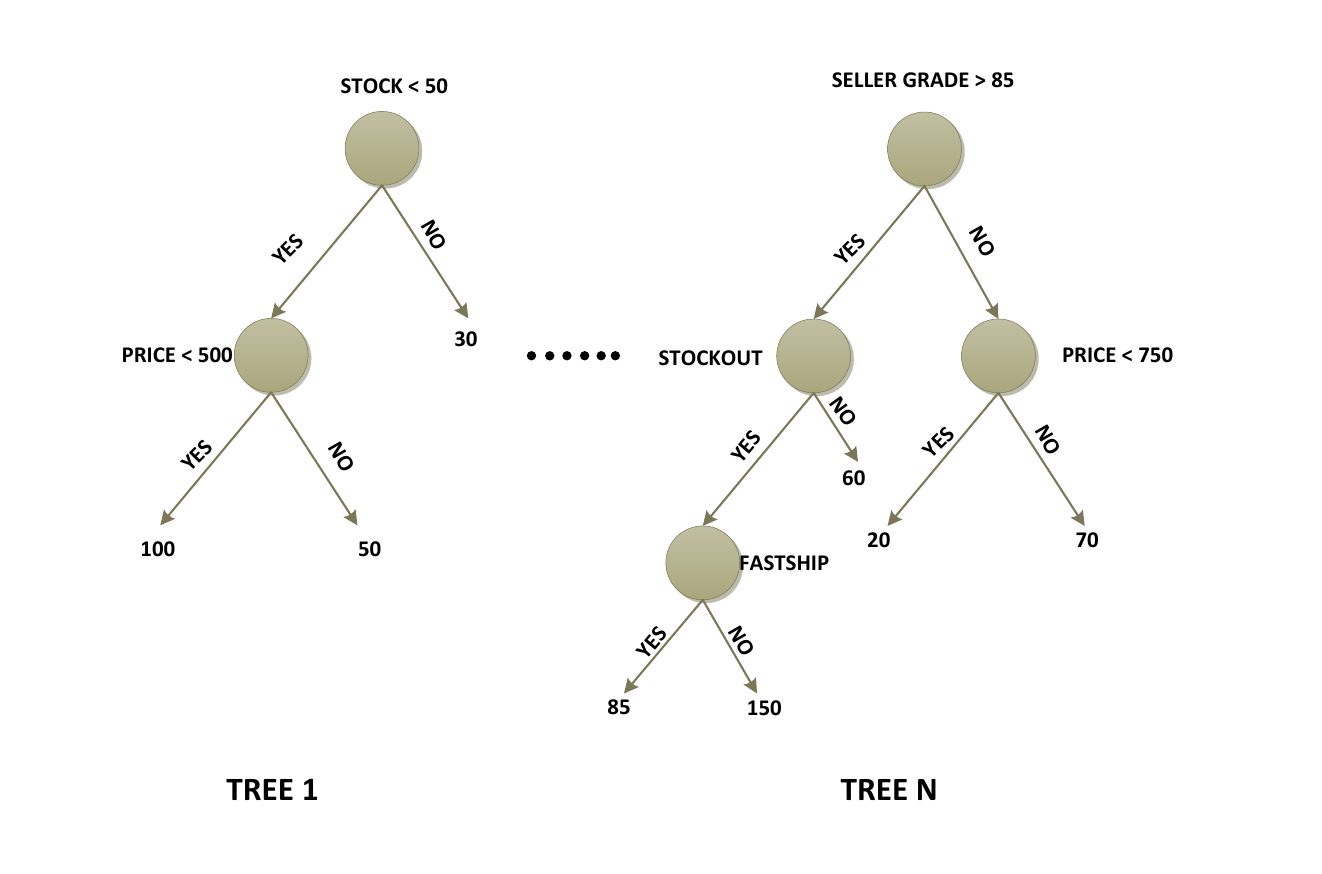}
	  \caption{Random Forest}
 	 \label{fig:fig4}
	\end{figure}

The probability of an example not being chosen in the dataset that has N samples is :

\begin{equation}
	Pr = 1 - \frac{1}{N}
\end{equation}
The probability of being in the test set for a sample is:
\begin{equation}
 Pr = \left(1 - \frac{1}{N}\right)^N \approx \exp^{-1} = 0.368
\end{equation}
Every tree has a different test set and this set consists of totally $\%63.2$ of data. Samples in the test set are called out-of-bag data. On the other hand, every tree has different features which are selected randomly. While selecting nodes in the tree, only a subset of the features are selected and the best one is chosen as separator node from this subset. Then this process continues recursively until a certain error rate is reached. Each tree is grown independently to reach the specified error rate. For instance, stock feature is chosen as the best separator node among the other randomly selected features, and likewise price feature is chosen as second best node for the first tree in Figure \ref{fig:fig4}. This tree is constructed with two nodes such as stock and price, whereas TREE N has four nodes and some of them are different than the TREE 1. 

Due to bootstrapping sampling method, there is no need to use cross-validation or separate datasets for training and testing. This process is done internally. In this project, minimum root mean squared error was achieved by using random forest with 20 trees in the first level.

\subsection{Gradient Boosted Trees}
	
	Gradient Boosted Trees (GBT) are ensemble learning of decision trees \cite{friedman2001greedy}. GBT are said to be the combination of gradient descent and boosting algorithms. Boosting methods aim at improving the performance of classification task by converting weak learners to strong ones. There are multiple boosting algorithms in literature \cite{oza2001experimental}, \cite{grabner2006line}, \cite{grabner2006real}, \cite{tutz2006generalized}. Adaboost is the first boosting algorithm proposed by Freund and Schapire \cite {freund1999short}. It works by weighting each sample in the dataset. Initially all samples are weighted equally likely and after each training iteration, misclassified samples are re-weighted more heavily. Boosting algorithms consist of many weak learners and use weighted summation of them. A weak learner can be defined as a learner that performs better than random guessing and it is used to compensate the shortcomings of existing weak learners. Gradient boosted trees uses gradient descent algorithm for the shortcomings of weak learners instead of using re-weighting mechanism. This algorithm is used to minimize the loss function (also called error function) by moving in the opposite direction of the gradient and finds a local minimum. In literature, there are several different loss functions such as Gaussian $L_2$, Laplace $L_1$, Binomial Loss functions etc \cite{natekin2013gradient}. Squared-error loss function, commonly used in many regression problems, is used in this project.

 Let $L(y_i, F(x_i))$ be the loss function where $y_i$ is actual output and $F(x_i)$ is the model we want to fit in. Our aim is to minimize $J = \sum_{i=1}^{N} (y_i - F(x_i))^2$ function.By using gradient descent algorithm,
 
\begin{equation} \label{eq:gds} 
	F(x_i) = F(x_i) - \alpha\frac{\delta J}{\delta F(x_i)}
\end{equation}

In Equation \ref{eq:gds}, $\alpha$ is learning rate that accelerates or decelerates the learning process. If the learning rate is very large then optimal or minimal point may be skipped. If the learning rate is too small, 
more iterations are required to find the minimum value of the loss function. While trees are constructed in parallel or independently in random forest ensemble learning, they are constructed sequential in gradient boosting. Once all trees have been trained, they are combined to give the final output.

	\section{EXPERIMENTAL RESULTS}

	\subsection{Dataset Definition}
	The data used in the experiments was provided from one of the most popular online e-commerce company in Turkey. First, standard preprocessing techniques are applied. Some of these techniques include filling in the missing values, removal of missing attributes when a major portion of the attribute values are missing and removal of irrelevant attributes. Each product/good has a timestamp which represents the date it is sold consisting of year, month, week and day information. A product can be sold several times within the same day from both same and different sellers. The demands or sales of a product are aggregated weekly. While the dataset contains 3575 instances and 17 attributes, only 1925 instances remained after the aggregation. Additionally, customers enter the company's website and choose a product they want. When they buy that product, this operation is inserted into a table as an instance, but if they give up to buy, this operation is also inserted into another table. We used this information to find the popularity of the product/good. For instance, product A is viewed 100 times and product B is viewed 55 times from both same and different users. It can be concluded that product A is more popular than product B. Before applying stacked generalization
method, outliers were removed, we only consider
the products where demand is less than 20. In this study, the parameters at the data preparation  stages are  determined by consulting with our contacts at the e-commerce company.
	\subsection{Evaluation Method}
We used Root Mean Squared Error (RMSE) to evaluate model performances. It is square root of the summation of differences between actual and predicted values. RMSE is frequently used in regression analysis. RMSE can be calculated as shown in Equation \ref{eq:rmse}. $\hat{\gamma_i}$'s are predicted and ${\gamma}$'s are actual values.
\begin{equation} \label{eq:rmse}
	RMSE = \sqrt{\frac{\sum_{i=1}^{n}({\hat{\gamma_i} - \gamma})^2}{n}}
\end{equation}

     \begin{figure}[b]
	\centering
 	 \includegraphics[width=\linewidth]			{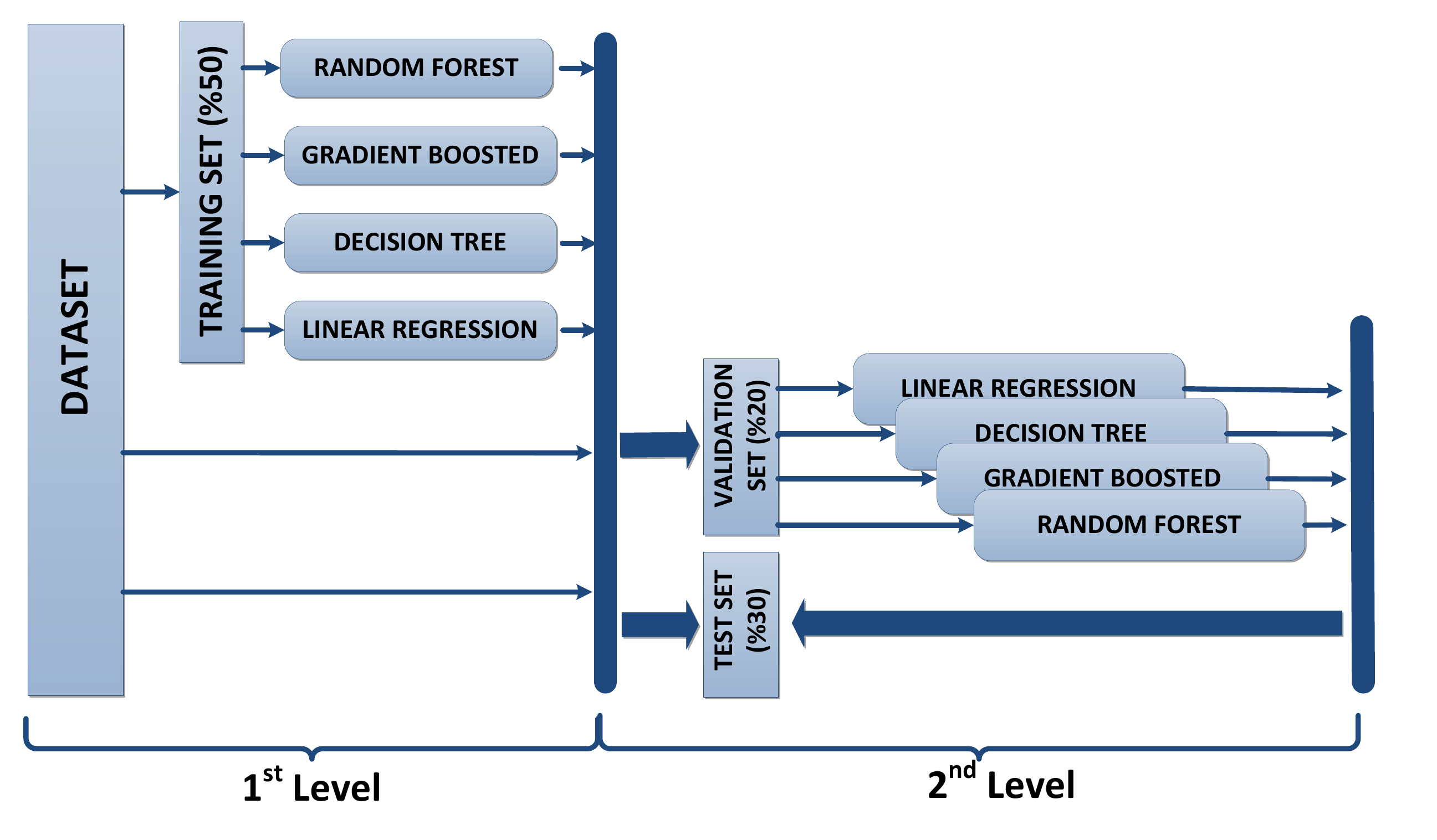}
	  \caption{Stacking Process}
 	 \label{fig:fig6}
	\end{figure}

\begin{figure*}
\begin{floatrow}[4]

\capbtabbox{%
 \begin{tabular}{|l|c|c|} 
 \hline

Model&RMSE \\ \hline 
DT&2.200\\ \hline
GBT&2.299\\ \hline
RF&2.120\\ \hline
LR&1.910\\ \hline
\end{tabular} 	
}{%
  \caption{The Results of Regressors at Level 2.
  \thinspace \thinspace \thinspace \enspace \enspace \enspace \enspace \enspace\enspace\enspace\enspace\enspace\enspace\enspace\enspace\enspace\enspace\enspace\enspace\enspace\enspace\enspace\enspace\enspace
    \label{Table1}}%
}

\capbtabbox{%
 \begin{tabular}{|l|c|c|}
 \hline
Model&RMSE \\ \hline 
DT&1.928\\ \hline
GBT&1.918\\ \hline
RF&1.865\\ \hline
LR&2.708\\ \hline
SG(LR)&1.864\\ \hline
\end{tabular} 
}{%
  \caption{The Best Results of Single Classifiers and SG.  \\ \hfill
   \label{Table2}
}%
}

\capbtabbox{%
 \begin{tabular}{|l|c|c|} 
 \hline
Model&RMSE \\ \hline 
GBT+DT&1,955		\\ \hline
GBT+LR&1,957	\\ \hline
LR+DT&1,962	\\ \hline
DT+RF&1,963		\\ \hline
GBT+RF&1,870		\\ \hline
LR+RF&1,927	\\ \hline
\end{tabular} 	
}{%
  \caption{SG with Binary Combination of Regressors.
   \\ \hfill
    \label{Table3}
  }%
}

\capbtabbox{%
 \begin{tabular}{|l|c|c|} 
 \hline
Model&RMSE \\ \hline 
DT+RF+GBT&1.894\\ \hline
RF+DT+LR&1.909\\ \hline
GBT+LR+DT&2.011\\ \hline
LR+RF+GBT&1.864\\ \hline
\end{tabular} 
}{%
  \caption{SG with Triple    Combination
of Regressors.  \\ \hfill \break
    \label{Table4}
}%
}

\end{floatrow}
\end{figure*}

We compared the results of the SG method with the results obtained by single classifiers. These classifiers include DT, GBT, RF and LR. Firstly, we split the data into training, validation and test sets (use $\%50$ of data for training, $\%20$ of data for validation and the remaining part for testing) and trained first level regressors using the training set. For SG, we applied 10-fold cross validation on the training set to get the best model of the first level regressors (except random forest ensemble model). After getting the first level regressor models, we used the validation set to create second level of the SG model. The single classifiers are trained on the combined training and test sets. The results of single classifiers and SG are evaluated using the test set in terms of RMSE. This process can be seen in Figure \ref{fig:fig6}.

\subsection{Result and Discussion}
	In this section, we evaluate the proposed model using RMSE evaluation method. After calculating RMSE for single classifiers and SG, we applied analysis of variances (ANOVA) test. It is generalized version of t-test to determine whether there are any statistically significant differences between the means of two or more unrelated groups. We use ANOVA test to show that predictions of the models are statistically different. The training set is divided randomly into 20 different subsets, so that no subset contains the whole training set. Using each of the different subsets and the validation set, the SG model is trained and evaluated on the test set. In the first level of the SG model, various combinations of the four algorithms are used.  We also conducted experiments with different machine learning algorithms in the second level of SG. For the single classifiers, the combination of training and tests is divided randomly into 20 subsets, and the same evaluation process is also repeated for these classifiers. We also run the proposed method with different combinations of the first level regressors. 
	
Table \ref{Table1} shows the the average of the RMSE results of the SG when using different learning methods in the second level. As can be seen from the table, LR outperforms the other learning methods. For this reason, in the remaining experiments, the results of the SG model is given when using LR in the second level. 
Table \ref{Table2} shows the best results of single classifiers and SG model  obtained from 20 runs. The SG model gives the best result when LR, RF and GBT are used in the first level. Table \ref{Table3} shows the results of binary combinations of the models. We found the minimum RMSE as $1.870$ by using GBT and RF together in the first level. After using binary combination of the models in the first level, we also created triple combination of them to specify the best combination. Table \ref{Table4} shows results of triple combination of models.

In ANOVA test, the null hypothesis rejected with $5\%$ significance level which shows that the predictions of RF and LR are significantly better than others in the first and second level respectively.

After concluding RF and LR are statistically different than other regressors at level 1 and 2 respectively, we applied t-test again with $\alpha = 0.05$ between RF in the first level and LR in the second level. Result of the t-test showed that LR in the second level is not statistically significantly different than RF.
	\section{CONCLUSION AND FUTURE WORK}
\noindent In this paper, we examine the problem of demand forecasting on an e-commerce web site. We proposed stacked generalization method consists of sub-level regressors. We have also tested results of single classifiers separately together with the general model. Experiments have shown that our approach predicts demand at least as good as single classifiers do, even better using much less training data (only $\%20$ of the dataset). We think that our approach will predict much better than other single classifiers when more data is used. Because of the difference is not statistically significant between the proposed model and random forest, the proposed method can be used to forecast demand due to its accuracy with less data. In the future, we will use the output of this project as part of price optimization problem which we are planning to work on.

\section*{\uppercase{Acknowledgements}}
\noindent The data used in this study was provided by n11 (www.n11.com). The authors are also thankful for the assistance rendered by \.{I}lker K\"{u}\c{c}\"{u}kil in the production of the data set.

\bibliographystyle{apalike}
{\small
\bibliography{demand}}

\vfill
\end{document}